\documentclass{article}
\usepackage{arxiv}
\usepackage{times}
\usepackage[numbers,sort&compress]{natbib}
\usepackage{hyperref}
\usepackage{url}
\usepackage{graphicx}
\usepackage{amsmath,amssymb}
\usepackage{xcolor}
\usepackage{subcaption}
\usepackage{multirow}
\usepackage{longtable}
\usepackage{booktabs}
\usepackage{placeins}
\usepackage{float}
\usepackage{etoolbox}
\usepackage{tikz}
\usetikzlibrary{positioning,fit,shapes.geometric,shapes.misc,
                arrows.meta,calc,backgrounds,shadows,
                decorations.pathreplacing}
\graphicspath{{figures/}}
\providecommand{\Description}[1]{}

\hypersetup{colorlinks=true,linkcolor=blue!60!black,citecolor=blue!60!black,urlcolor=blue!60!black}

\title{\normalfont RECITE: Revising Eligibility Criteria Incorporating Textual Evidence}

\author{
  \textbf{Russell Ro}$^{1,2}$, \textbf{Reza Abbasi-Asl}$^{1}$\thanks{Corresponding author: \texttt{reza.abbasiasl@ucsf.edu}} \\[6pt]
  $^{1}$University of California, San Francisco \quad
  $^{2}$University of California, Berkeley \\[4pt]
}

\date{}

\renewcommand{\shorttitle}{RECITE: Revising Eligibility Criteria Incorporating Textual Evidence}

\begin{document}

\maketitle

\begin{abstract}
Overly restrictive eligibility criteria contribute to widespread enrollment failure in clinical trials, yet published recommendations for how specific criteria should be broadened remain disconnected from trial design at scale. We introduce RECITE (Revising Eligibility Criteria Incorporating Textual Evidence), a framework for automated discovery, extraction, and implementation of evidence-based eligibility criteria revisions. RECITE uses protocol amendment documents from ClinicalTrials.gov to construct a dataset of 11{,}913 curated instances across 5{,}735 trials, each pairing original criteria with documentary evidence and target revisions. We benchmark fifteen LLM configurations under truncated-to-context-window and retrieval-augmented generation conditions, on a 3{,}116-instance sample under an LLM-as-judge protocol validated against ten domain experts (judge vs.\ human quadratic-weighted Cohen's $\kappa_{qw} = 0.694 \pm 0.096$). The best open-weight model, Gemma~2 9B, achieves $\kappa_{qw}$-aligned ordinal score $3.0 \pm 0.9$ on a 0--4 scale (normalized 0.742) and the best closed-weight model, GPT-4o, reaches $3.0 \pm 0.8$ (normalized 0.745). Using an agentic literature discovery system, we then identified 44 paper--trial matches across five publications providing trial-design and EC-modification evidence. Our findings suggest that existing clinical knowledge can be operationalized into actionable trial design recommendations.
\end{abstract}


\vspace{6pt}
\noindent An earlier version of this work appears in the Proceedings of the 32nd ACM SIGKDD Conference on Knowledge Discovery and Data Mining (KDD '26), \url{https://doi.org/10.1145/3770855.3819027}.

\vspace{6pt}
\begingroup\small\noindent\raggedright\textbf{Resource Availability:}
The source code and benchmark data are publicly available at \url{https://doi.org/10.5281/zenodo.20487342}. The source repository is hosted at \url{https://github.com/abbasilab/RECITE}.
\endgroup


\section{Introduction}
\label{sec:introduction}

\begin{figure}[!htbp]
\centering
\begin{tikzpicture}[
  stage/.style 2 args={
    rectangle, rounded corners=3pt,
    draw=#1, fill=#2,
    thick, text width=0.82\columnwidth,
    minimum height=0.95cm, inner sep=5pt,
    font=\small, align=left,
  },
  arrow/.style={-{Stealth[scale=0.8]}, thick, line width=0.7pt, #1},
  arrow/.default={},
  note/.style={font=\tiny\itshape, text=gray!60!black},
]

  \node[stage={blue!60!black}{blue!6}] (s1) {%
    \textbf{1.\ Benchmark Construction} {\tiny(\S\ref{sec:benchmark})}\\[-1pt]
    {\footnotesize 568K CTG trials $\to$ EC change detection $\to$ protocol PDF extraction $\to$ \textbf{11{,}913 instances}}%
  };

  \node[stage={orange!70!black}{orange!6}, below=0.45cm of s1] (s2) {%
    \textbf{2.\ Model Evaluation} {\tiny(\S\ref{sec:benchmark})}\\[-1pt]
    {\footnotesize 8 LLMs + full-doc/RAG + LLM-as-judge $\to$ \textbf{85.8\% equiv.}}%
  };

  \node[stage={teal!70!black}{teal!6}, below=0.45cm of s2] (s3) {%
    \textbf{3.\ Evidence Discovery} {\tiny(\S\ref{sec:discovery})}\\[-1pt]
    {\footnotesize PubMed/Semantic Scholar crawl $\to$ directive + accrual extraction $\to$ author validation}%
  };

  \node[stage={green!50!black}{green!6}, below=0.45cm of s3] (s4) {%
    \textbf{4.\ Accrual Impact} {\tiny(\S\ref{sec:applications})}\\[-1pt]
    {\footnotesize Paper--trial matching + accrual impact estimation $\to$ \textbf{25--143\% est.\ gain}}%
  };

  \draw[arrow={blue!50!black}]
    (s1.south) -- node[right, note]{select best model} (s2.north);
  \draw[arrow={orange!50!black}]
    (s2.south) -- node[right, note]{deploy model} (s3.north);
  \draw[arrow={teal!50!black}]
    (s3.south) -- node[right, note]{match to trials} (s4.north);

\end{tikzpicture}
\caption{RECITE framework overview.
Four stages: benchmark construction from ClinicalTrials.gov protocols
(\S\ref{sec:benchmark}),
model evaluation via LLM-as-judge,
evidence discovery from published literature
(\S\ref{sec:discovery}),
and estimated accrual impact
(\S\ref{sec:applications}).}
\Description{Vertical four-stage pipeline diagram: benchmark construction,
model evaluation, evidence discovery, and accrual impact estimation.}
\label{fig:framework_overview}
\end{figure}
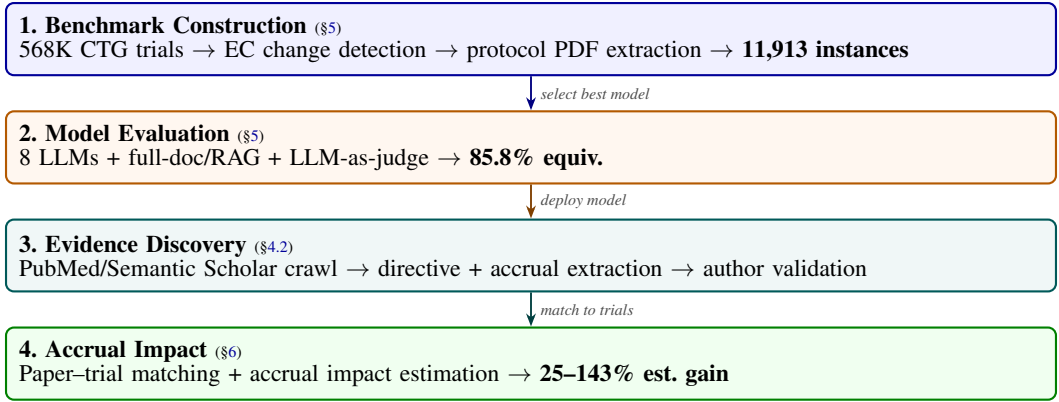

Clinical trials are the primary mechanism for evaluating medical interventions, yet enrollment remains a persistent bottleneck. Only 31\% of UK Medical Research Council trials met their original enrollment goals~\cite{bower2009recruitment}, one-third of publicly funded trials required time extensions due to recruitment shortfalls~\cite{campbell2007steps}, and 25\% of cancer trials failed to enroll a sufficient number of participants~\cite{feller2015cancer}. Insufficient accrual is the single largest cause of trial termination, accounting for 36.7\% of all stoppages in a recent analysis of 28,561 discontinued trials~\cite{razuvayevskaya2024genetic}. When enrollment shortfalls trigger protocol amendments, the costs compound: 76\% of protocols undergo at least one amendment, and each Phase~III amendment costs a median of \$535,000~\cite{getz2016amendments, getz2024benchmarks}. Overly restrictive eligibility criteria are widely recognized as a contributing factor~\cite{liu2021trial}.

A growing body of work already articulates explicit recommendations for how specific eligibility criteria should be loosened. Governing body directives from the American Society of Clinical Oncology and Friends of Cancer Research (ASCO-Friends) recommend inclusion of patients with controlled brain metastases, prior malignancies, and reduced creatinine clearance thresholds~\cite{kim2017asco}. The U.S. Food and Drug Administration has issued cross-cutting guidance encouraging sponsors to broaden criteria, justify each exclusion, and include medically complex participants~\cite{fda2020diversity}. Published research provides disease-specific evidence across therapeutic areas including oncology~\cite{kaur2024broadening, liu2021trial}, cardiology~\cite{zannad2024cardiology}, neurology~\cite{langbaum2023alzheimer}, psychiatry~\cite{zimmerman2019depression}, rheumatology~\cite{dyball2021lupus}, and nephrology~\cite{colombijn2024ckd}, supported by consensus directives from KDIGO~\cite{baigent2017kdigo} and the Global Cardiovascular Clinical Trialists Forum~\cite{filbey2023representativeness}.

Despite this wealth of guidance, no existing system automates the discovery, extraction, and implementation of such recommendations across trials at scale. Individual trialists may consult specific publications when designing or amending protocols, but systematically connecting the large body of published recommendations to active trial designs remains an unsolved technical problem. The gap is not one of clinical knowledge (the recommendations exist) but of infrastructure: the retrieval and matching between two large bodies of text, published evidence and registered trials, exceeds what manual processes can achieve.

Recent computational work has addressed adjacent problems, including amendment prediction~\cite{das2026amend} and criteria generation from trial descriptions~\cite{lekuthai2025ecraft, wang2023autotrial}, but none automates evidence-grounded revision (Section~\ref{sec:related}).

To address this gap, we introduce RECITE (\textbf{R}evising \textbf{E}ligibility \textbf{C}riteria \textbf{I}ncorporating \textbf{T}extual \textbf{E}vidence), a framework for evidence-guided eligibility criteria revision. Given original eligibility criteria and supporting documentary evidence, whether from protocol amendment documents, published research papers, or clinical guidelines, RECITE extracts explicit modification directives and generates revised criteria with cited provenance. The framework operates across diverse evidence sources and handles the variation in structure, length, and style typical of real-world clinical documents.


\definecolor{amber}{RGB}{255,191,0}
\begin{figure}[!htbp]
\setlength{\columnwidth}{0.6\textwidth}
\centering

\begin{subfigure}[b]{\columnwidth}
\centering
\resizebox{\columnwidth}{!}{%
\begin{tikzpicture}[
  every node/.style={font=\footnotesize, align=center},
  node distance=0.4cm and 0.6cm,
  data node/.style={
    cylinder, shape border rotate=90, aspect=0.25,
    draw=blue!70, fill=blue!15, thick,
    minimum width=1.5cm, minimum height=1.1cm, inner sep=3pt,
  },
  standard process/.style={
    rectangle, rounded corners=2pt,
    draw=amber!80!black, fill=amber!18!yellow!12, thick,
    inner sep=3pt, minimum height=1.1cm,
  },
  decision node/.style={
    diamond, draw=purple!70, fill=purple!10, thick,
    inner sep=1pt, aspect=2.2,
  },
  output node/.style={
    rounded rectangle,
    draw=green!60!black, fill=green!12, thick,
    inner sep=4pt, minimum height=1.1cm,
  },
  discard node/.style={
    rectangle, rounded corners=2pt,
    draw=red!30, fill=red!6, thick,
    inner sep=2pt, minimum height=0.9cm, font=\small\color{gray!70},
  },
  arrow/.style={-{Stealth[scale=0.9]}, thick, line width=0.8pt},
  subgraph box/.style={
    draw=gray!50, rounded corners=4pt, inner sep=8pt, thick,
  },
]
  \node[data node, text width=1.7cm]                         (xml)  {ClinicalTrials\\{.gov} Bulk XML};
  \node[standard process, right=of xml, text width=1.8cm]    (load) {Load 567{,}815\\NCT IDs};
  \node[decision node, right=1.0cm of load, text width=2.2cm](dec1) {$\geq 2$ versions\\w/ EC change?};
  \node[output node, right=1.0cm of dec1, text width=1.6cm]  (out1) {59{,}996\\Trials};
  \node[discard node, below=0.5cm of dec1]                    (disc1){507K\\filtered};

  \draw[arrow] (xml)  -- (load);
  \draw[arrow] (load) -- (dec1);
  \draw[arrow] (dec1) -- node[above, font=\tiny\itshape, yshift=4pt]{Yes 10.6\%} (out1);
  \draw[arrow, gray!50] (dec1) -- node[right, font=\tiny\itshape]{No} (disc1);

  \begin{scope}[on background layer]
    \node[subgraph box, fill=blue!4,
          fit=(xml)(load)(dec1)(out1)(disc1)] (bgA) {};
  \end{scope}
  \node[anchor=south west, font=\footnotesize\bfseries] at (bgA.south west) {(a)};
\end{tikzpicture}%
}
\label{fig:benchmark_discovery}
\end{subfigure}

\vspace{0.5em}

\begin{subfigure}[b]{\columnwidth}
\centering
\resizebox{\columnwidth}{!}{%
\begin{tikzpicture}[
  every node/.style={font=\footnotesize, align=center},
  node distance=0.4cm and 0.6cm,
  standard process/.style={
    rectangle, rounded corners=2pt,
    draw=amber!80!black, fill=amber!18!yellow!12, thick,
    inner sep=3pt, minimum height=1.1cm,
  },
  decision node/.style={
    diamond, draw=purple!70, fill=purple!10, thick,
    inner sep=1pt, aspect=2.2,
  },
  output node/.style={
    rounded rectangle,
    draw=green!60!black, fill=green!12, thick,
    inner sep=4pt, minimum height=1.1cm,
  },
  discard node/.style={
    rectangle, rounded corners=2pt,
    draw=red!30, fill=red!6, thick,
    inner sep=2pt, minimum height=0.9cm, font=\small\color{gray!70},
  },
  arrow/.style={-{Stealth[scale=0.9]}, thick, line width=0.8pt},
  subgraph box/.style={
    draw=gray!50, rounded corners=4pt, inner sep=8pt, thick,
  },
]
  \node[output node, text width=1.5cm]                         (in2)  {59{,}996\\Trials};
  \node[standard process, right=of in2, text width=2.0cm]      (dl)   {Download\\Versions\\(269K records)};
  \node[decision node, right=1.0cm of dl, text width=1.8cm]   (dec2) {EC text\\changed?};
  \node[output node, right=1.0cm of dec2, text width=2.2cm]   (out2) {85{,}083 changes\\(49{,}369 trials)};
  \node[discard node, below=0.5cm of dec2]                     (disc2){Metadata\\only};

  \draw[arrow] (in2)  -- (dl);
  \draw[arrow] (dl)   -- (dec2);
  \draw[arrow] (dec2) -- node[above, font=\tiny\itshape]{Yes} (out2);
  \draw[arrow, gray!50] (dec2) -- node[right, font=\tiny\itshape]{No} (disc2);

  \begin{scope}[on background layer]
    \node[subgraph box, fill=orange!4,
          fit=(in2)(dl)(dec2)(out2)(disc2)] (bgB) {};
  \end{scope}
  \node[anchor=south west, font=\footnotesize\bfseries] at (bgB.south west) {(b)};
\end{tikzpicture}%
}
\label{fig:benchmark_ec}
\end{subfigure}

\vspace{0.5em}

\begin{subfigure}[b]{\columnwidth}
\centering
\resizebox{\columnwidth}{!}{%
\begin{tikzpicture}[
  every node/.style={font=\footnotesize, align=center},
  node distance=0.4cm and 0.6cm,
  standard process/.style={
    rectangle, rounded corners=2pt,
    draw=amber!80!black, fill=amber!18!yellow!12, thick,
    inner sep=3pt, minimum height=1.1cm,
  },
  decision node/.style={
    diamond, draw=purple!70, fill=purple!10, thick,
    inner sep=1pt, aspect=2.2,
  },
  output node/.style={
    rounded rectangle,
    draw=green!60!black, fill=green!12, thick,
    inner sep=4pt, minimum height=1.1cm,
  },
  discard node/.style={
    rectangle, rounded corners=2pt,
    draw=red!30, fill=red!6, thick,
    inner sep=2pt, minimum height=0.9cm, font=\small\color{gray!70},
  },
  deliverable/.style={
    rectangle, rounded corners=2pt,
    draw=green!60!black, fill=green!18, very thick,
    inner sep=4pt, minimum height=1.1cm,
    font=\small\bfseries,
  },
  arrow/.style={-{Stealth[scale=0.9]}, thick, line width=0.8pt},
  subgraph box/.style={
    draw=gray!50, rounded corners=4pt, inner sep=8pt, thick,
  },
]
  \node[output node, text width=1.8cm]                         (in3)  {49{,}369\\Trials};
  \node[decision node, right=1.0cm of in3, text width=1.6cm]  (dec3) {Protocol\\PDF?};
  \node[output node, right=1.2cm of dec3, text width=1.8cm]   (pdf)  {$\sim$3{,}500\\w/ evidence};
  \node[discard node, below=0.5cm of dec3]                     (disc3){$\sim$46K\\no protocol};
  \node[standard process, right=0.8cm of pdf, text width=1.8cm] (join) {JOIN changes\\+ evidence};

  \node[output node, below=0.9cm of join, text width=2.0cm]   (inst) {11{,}913 RECITE\\Instances};
  \node[deliverable, left=0.8cm of inst, text width=2.0cm]    (samp) {Sample\\3{,}116};

  \draw[arrow] (in3)  -- (dec3);
  \draw[arrow] (dec3) -- node[above, font=\tiny\itshape]{Yes ($\sim$7\%)} (pdf);
  \draw[arrow, gray!50] (dec3) -- node[right, font=\tiny\itshape]{No} (disc3);
  \draw[arrow] (pdf)  -- (join);
  \draw[arrow] (join) -- (inst);
  \draw[arrow] (inst) -- (samp);

  \begin{scope}[on background layer]
    \node[subgraph box, fill=teal!4,
          fit=(in3)(dec3)(pdf)(disc3)(join)(inst)(samp)] (bgC) {};
  \end{scope}
  \node[anchor=south west, font=\footnotesize\bfseries] at (bgC.south west) {(c)};
\end{tikzpicture}%
}
\label{fig:benchmark_assembly}
\end{subfigure}

\caption{RECITE benchmark construction pipeline.
\textbf{(a)}~Discovery and version filtering.
\textbf{(b)}~EC change detection across consecutive protocol versions.
\textbf{(c)}~Evidence extraction from protocol PDFs and benchmark assembly.}
\Description{Three-part funnel diagram showing benchmark construction from 567K trials
through filtering stages to 3{,}116 evaluation instances.}
\label{fig:benchmark_pipeline}
\end{figure}
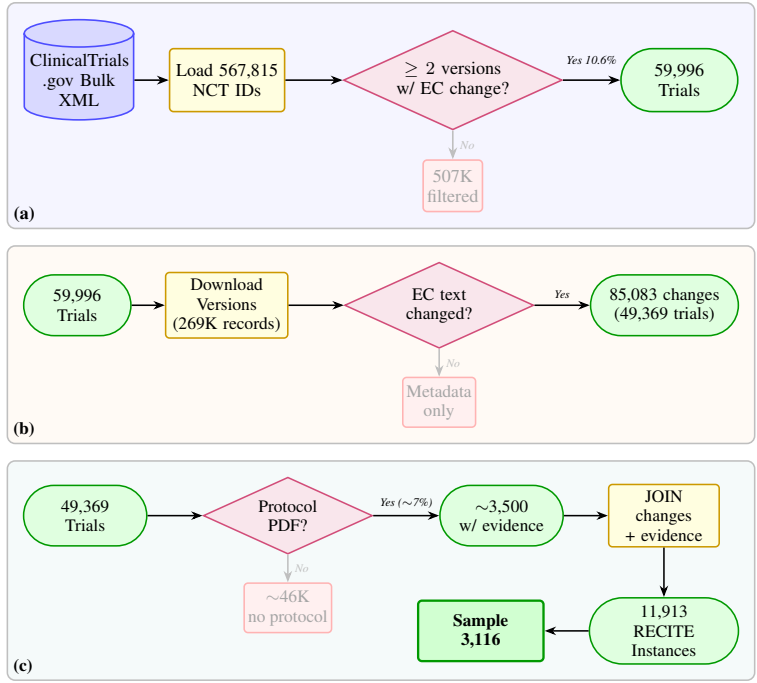

Our contributions are as follows:
\begin{enumerate}
    \item We formulate the task of \textit{evidence-guided eligibility criteria revision}, defining it over diverse documentary sources including protocol amendment documents, published literature, and clinical guidelines.
    \item We develop an agentic literature discovery system that crawls PubMed, matches papers to active trials, and extracts actionable modification directives.
    \item We construct the RECITE benchmark from 567,815 trials on ClinicalTrials.gov, filtering through version histories and protocol evidence to yield 11,913 instances across 5,735 unique trials. We evaluate on a sample of 3,116 instances with original criteria, grounding evidence from protocol amendment documents, and target revisions.
    \item We demonstrate real-world applicability: 44 paper--trial matches across five publications surface evidence-grounded EC-broadening directives. The results show that existing clinical knowledge can be operationalized into evidence-based trial design (with directional, not numeric, accrual claims).
\end{enumerate}

\section{Background and Related Work}
\label{sec:related}

\subsection{Eligibility Criteria Design and the Case for Liberalization}

Eligibility criteria balance participant safety with scientific validity in clinical trials. Evidence across therapeutic areas suggests that criteria more restrictive than necessary can reduce enrollment, limit demographic diversity, and narrow the generalizability of findings. Oncology has produced the largest evidence base for reform. The ASCO-Friends consensus statement provides concrete recommendations for broadening criteria related to brain metastases, prior malignancies, and organ function thresholds~\cite{kim2017asco}, while Trial Pathfinder demonstrated that data-driven relaxation of common exclusions does not compromise hazard ratios~\cite{liu2021trial}. Real-world data analyses have validated these recommendations, showing that applying ASCO-Friends expanded criteria can increase the eligible pool by 78\% or more~\cite{kaur2024broadening}. The FDA's 2020 guidance provides cross-cutting regulatory encouragement for sponsors to broaden criteria and justify each exclusion~\cite{fda2020diversity}.

This pattern extends beyond oncology: restrictive criteria exclude 63\% of lupus patients~\cite{dyball2021lupus}, 86\% of depressed outpatients from antidepressant trials~\cite{zimmerman2019depression}, and up to 88\% of candidates in preclinical Alzheimer's trials~\cite{langbaum2023alzheimer}. Franzen et al.~\cite{franzen2022diversity} show that common Alzheimer's exclusions reduce racial and ethnic diversity in enrolled populations, while eligibility complexity in dementia trials has increased over time despite calls for simplification~\cite{mitchell2024adrd}. The proportion of cardiovascular trials excluding chronic kidney disease patients has risen from 66\% to 79\% over twenty years~\cite{colombijn2024ckd}. Governing bodies have responded with directives from KDIGO~\cite{baigent2017kdigo}, the Global Cardiovascular Clinical Trialists Forum~\cite{filbey2023representativeness}, and ICH~E17~\cite{ich2017e17}. MDS-specific guidance has called for reducing eligibility barriers in hematologic malignancy trials~\cite{mds2025reducing}. Taken together, these are explicit, actionable recommendations distributed across governing bodies, academic literature, and institutional protocols.


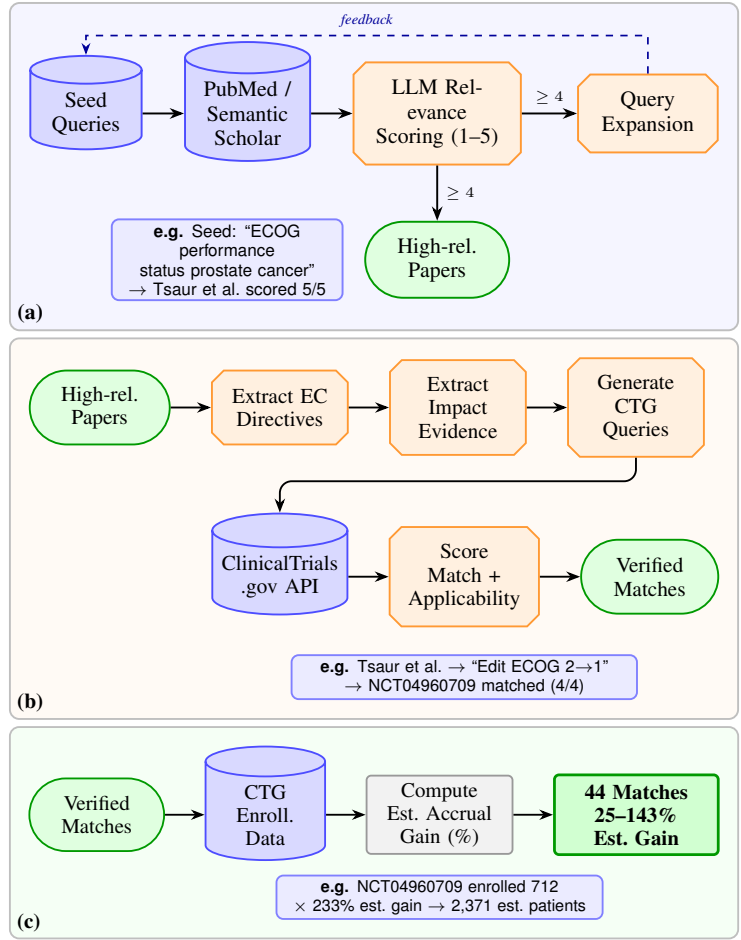
\begin{figure}[!htbp]
\setlength{\columnwidth}{0.75\textwidth}
\centering
\scalebox{0.78}{\begin{minipage}{\columnwidth}
\centering

\begin{subfigure}[b]{\columnwidth}
\centering
\resizebox{\columnwidth}{!}{%
\begin{tikzpicture}[
  every node/.style={font=\footnotesize, align=center},
  node distance=0.4cm and 0.6cm,
  data node/.style={
    cylinder, shape border rotate=90, aspect=0.25,
    draw=blue!70, fill=blue!15, thick,
    minimum width=1.5cm, minimum height=1.1cm, inner sep=3pt,
  },
  llm process/.style={
    chamfered rectangle, chamfered rectangle xsep=2pt,
    draw=orange!80, fill=orange!12, thick,
    inner sep=4pt, minimum height=1.1cm,
  },
  standard process/.style={
    rectangle, rounded corners=2pt,
    draw=gray!80, fill=gray!10, thick,
    inner sep=3pt, minimum height=1.1cm,
  },
  output node/.style={
    rounded rectangle,
    draw=green!60!black, fill=green!12, thick,
    inner sep=4pt, minimum height=1.1cm,
  },
  discard node/.style={
    rectangle, rounded corners=2pt,
    draw=red!30, fill=red!6, thick,
    inner sep=2pt, minimum height=0.9cm, font=\small\color{gray!70},
  },
  arrow/.style={-{Stealth[scale=0.9]}, thick, line width=0.8pt},
  feedback/.style={-{Stealth[scale=0.8]}, thick, line width=0.7pt, dashed},
  subgraph box/.style={
    draw=gray!50, rounded corners=4pt, inner sep=8pt, inner ysep=12pt, thick,
  },
  example box/.style={
    rectangle, rounded corners=3pt, draw=blue!50, thick,
    fill=blue!8, inner sep=2pt, font=\sffamily\scriptsize, text width=2.8cm,
  },
]
  \node[data node, text width=1.4cm]                            (seed)  {Seed\\Queries};
  \node[data node, right=of seed, text width=1.6cm]             (apis)  {PubMed /\\Semantic\\Scholar};
  \node[llm process, right=of apis, text width=2.0cm]           (score) {LLM Relevance\\Scoring (1--5)};
  \node[llm process, right=0.8cm of score, text width=1.6cm]    (expand){Query\\Expansion};
  \node[output node, below=0.8cm of score, text width=1.6cm] (keep) {High-rel.\\Papers};

  \draw[arrow] (seed)  -- (apis);
  \draw[arrow] (apis)  -- (score);
  \draw[arrow] (score) -- node[above, font=\tiny\itshape]{$\geq 4$} (expand);
  \draw[feedback, blue!60!black] (expand.north) -- ++(0,0.55) -| node[pos=0.25, above, font=\tiny\itshape]{feedback} (seed.north);
  \draw[arrow] (score) -- node[right, font=\tiny\itshape]{$\geq 4$} (keep);

  \node[example box, text width=3.2cm, left=0.3cm of keep] (exa)
    {\textbf{e.g.} Seed: ``ECOG performance\\status prostate cancer''\\$\rightarrow$ Tsaur et al.\ scored 5/5};

  \coordinate (apex) at ($(expand.north) + (0,0.6)$);
  \begin{scope}[on background layer]
    \node[subgraph box, fill=blue!4,
          fit=(seed)(apis)(score)(expand)(keep)(exa)(apex)] (bgA) {};
  \end{scope}
  \node[anchor=south west, font=\footnotesize\bfseries] at (bgA.south west) {(a)};
\end{tikzpicture}%
}
\label{fig:crawl}
\end{subfigure}

\vspace{0.2em}

\begin{subfigure}[b]{\columnwidth}
\centering
\resizebox{\columnwidth}{!}{%
\begin{tikzpicture}[
  every node/.style={font=\footnotesize, align=center},
  node distance=0.4cm and 0.6cm,
  data node/.style={
    cylinder, shape border rotate=90, aspect=0.25,
    draw=blue!70, fill=blue!15, thick,
    minimum width=1.5cm, minimum height=1.1cm, inner sep=3pt,
  },
  llm process/.style={
    chamfered rectangle, chamfered rectangle xsep=2pt,
    draw=orange!80, fill=orange!12, thick,
    inner sep=4pt, minimum height=1.1cm,
  },
  standard process/.style={
    rectangle, rounded corners=2pt,
    draw=gray!80, fill=gray!10, thick,
    inner sep=3pt, minimum height=1.1cm,
  },
  output node/.style={
    rounded rectangle,
    draw=green!60!black, fill=green!12, thick,
    inner sep=4pt, minimum height=1.1cm,
  },
  arrow/.style={-{Stealth[scale=0.9]}, thick, line width=0.8pt},
  subgraph box/.style={
    draw=gray!50, rounded corners=4pt, inner sep=8pt, thick,
  },
  example box/.style={
    rectangle, rounded corners=3pt, draw=blue!50, thick,
    fill=blue!8, inner sep=2pt, font=\sffamily\scriptsize, text width=2.8cm,
  },
]
  \node[output node, text width=1.6cm]                          (in4)  {High-rel.\\Papers};
  \node[llm process, right=of in4, text width=1.6cm]            (dir)  {Extract EC\\Directives};
  \node[llm process, right=of dir, text width=1.6cm]            (imp)  {Extract Impact\\Evidence};
  \node[llm process, right=of imp, text width=1.6cm]            (qgen) {Generate CTG\\Queries};

  \node[data node, below=1.0cm of dir, text width=1.8cm]        (ctg)  {ClinicalTrials\\{.gov} API};
  \node[llm process, right=of ctg, text width=1.8cm]            (eval) {Score Match +\\Applicability};
  \node[output node, right=of eval, text width=1.6cm]           (out4) {Verified\\Matches};

  \draw[arrow] (in4)  -- (dir);
  \draw[arrow] (dir)  -- (imp);
  \draw[arrow] (imp)  -- (qgen);
  \draw[arrow, rounded corners=5pt] (qgen.south) -- ++(0,-0.35) -| (ctg.north);
  \draw[arrow] (ctg)  -- (eval);
  \draw[arrow] (eval) -- (out4);

  \node[example box, text width=5.0cm, below=0.4cm of eval] (exb)
    {\textbf{e.g.} Tsaur et al.\ $\rightarrow$ ``Edit ECOG 2$\rightarrow$1''\\
     $\rightarrow$ NCT04960709 matched (4/4)};

  \begin{scope}[on background layer]
    \node[subgraph box, fill=orange!4,
          fit=(in4)(dir)(imp)(qgen)(ctg)(eval)(out4)(exb)] (bgB) {};
  \end{scope}
  \node[anchor=south west, font=\footnotesize\bfseries] at (bgB.south west) {(b)};
\end{tikzpicture}%
}
\label{fig:screen_match}
\end{subfigure}

\vspace{0.2em}

\begin{subfigure}[b]{\columnwidth}
\centering
\resizebox{\columnwidth}{!}{%
\begin{tikzpicture}[
  every node/.style={font=\footnotesize, align=center},
  node distance=0.4cm and 0.6cm,
  data node/.style={
    cylinder, shape border rotate=90, aspect=0.25,
    draw=blue!70, fill=blue!15, thick,
    minimum width=1.5cm, minimum height=1.1cm, inner sep=3pt,
  },
  standard process/.style={
    rectangle, rounded corners=2pt,
    draw=gray!80, fill=gray!10, thick,
    inner sep=3pt, minimum height=1.1cm,
  },
  output node/.style={
    rounded rectangle,
    draw=green!60!black, fill=green!12, thick,
    inner sep=4pt, minimum height=1.1cm,
  },
  deliverable/.style={
    rectangle, rounded corners=2pt,
    draw=green!60!black, fill=green!18, very thick,
    inner sep=4pt, minimum height=1.1cm,
    font=\small\bfseries,
  },
  arrow/.style={-{Stealth[scale=0.9]}, thick, line width=0.8pt},
  subgraph box/.style={
    draw=gray!50, rounded corners=4pt, inner sep=8pt, thick,
  },
  example box/.style={
    rectangle, rounded corners=3pt, draw=blue!50, thick,
    fill=blue!8, inner sep=2pt, font=\sffamily\scriptsize, text width=2.8cm,
  },
]
  \node[output node, text width=1.6cm]                          (in5)  {Verified\\Matches};
  \node[data node, right=of in5, text width=1.6cm]              (enrl) {CTG Enroll.\\Data};
  \node[standard process, right=0.6cm of enrl, text width=2.0cm] (calc) {Compute\\Est.\ Accrual\\Gain (\%)};
  \node[deliverable, right=0.6cm of calc, text width=2.2cm]     (out5) {44 Matches\\25--143\%\\Est.\ Gain};

  \draw[arrow] (in5) -- (enrl);
  \draw[arrow] (enrl) -- (calc);
  \draw[arrow] (calc) -- (out5);

  \node[example box, text width=4.8cm, below=0.3cm of calc] (exc)
    {\textbf{e.g.} NCT04960709 enrolled 712\\
     $\times$ 233\% est.\ gain $\rightarrow$ 2{,}371 est.\ patients};

  \begin{scope}[on background layer]
    \node[subgraph box, fill=green!4,
          fit=(in5)(enrl)(calc)(out5)(exc)] (bgC) {};
  \end{scope}
  \node[anchor=south west, font=\footnotesize\bfseries] at (bgC.south west) {(c)};
\end{tikzpicture}%
}
\label{fig:accrual}
\end{subfigure}

\end{minipage}}
\caption{Agentic literature discovery system.
(a)~Iterative crawl: seed queries retrieve papers from PubMed and Semantic Scholar; an LLM scores relevance (1--5) and high-scoring papers expand the query set.
(b)~Screening and matching: directives and impact evidence are extracted from each paper, CTG API queries are generated, and candidate (paper, trial) pairs are scored.
(c)~Accrual estimation: verified matches are joined with CTG enrollment data to compute estimated gains (25--143\%).}
\Description{Three-part flow diagram of the agentic discovery pipeline.}
\label{fig:crawling}
\end{figure}

\subsection{Predicting and Implementing Criteria Changes}

AMEND++~\cite{das2026amend} introduces eligibility criteria amendment prediction as a supervised task, using Change-Aware Masked Language Modeling on historical version diffs spanning 161K trials. Their benchmark provides a resource for studying EC instability over time. RECITE shares the ClinicalTrials.gov data source but addresses a different goal: rather than predicting \textit{whether} amendments will occur, RECITE discovers evidence specifying \textit{what} changes should be made and generates the revised criteria accordingly. The two approaches are complementary: AMEND++ identifies trials at high amendment risk, while RECITE supplies evidence-grounded revisions. TrialPanorama~\cite{wang2025trialpanorama} provides a large-scale trial database with benchmark tasks but does not address evidence-guided revision. Kim et al.~\cite{liu2021covid} demonstrate data-driven eligibility criteria optimization for COVID-19 trials using real-world data, an approach complementary to our evidence-guided revision from published literature.

\subsection{AI for Clinical Trial Design and Literature Mining}

Generative approaches produce eligibility criteria from trial-level metadata. EC-RAFT~\cite{lekuthai2025ecraft} uses retrieval-augmented fine-tuning to generate criteria from trial descriptions, while AutoTrial~\cite{wang2023autotrial} prompts language models for protocol generation. Embedding-based methods have also addressed trial-level representation; for example, Trial2Vec~\cite{wang2022trial2vec} learns self-supervised trial document embeddings for similarity search, and Bornet et al.~\cite{bettiol2025eclusters} cluster eligibility criteria using sentence-level LLM embeddings, achieving 95\% of generative-LLM ROUGE scores with compressed cluster representations. These approaches capture distributional patterns across existing criteria. Downstream tasks such as patient--trial matching~\cite{jin2024trialgpt} and criteria-to-query translation~\cite{park2024criteria2query} would benefit from improved criteria. RECITE builds on retrieval-augmented generation~\cite{lewis2020rag} and dense passage retrieval~\cite{karpukhin2020dpr} to connect published recommendations to trial-level eligibility criteria.

\section{Problem Formulation}
\label{sec:formulation}

The task of \textit{evidence-guided eligibility criteria revision} is defined as follows. Given a set of original eligibility criteria $E_{\text{orig}}$ for a clinical trial and a documentary evidence source $D$ (which may be a protocol amendment document, a published research paper, a clinical guideline, or any other text containing recommendations for criteria modification), the goal is to produce revised eligibility criteria $E_{\text{rev}}$ with provenance. Specifically, a provenance mapping $P$ links each revision to specific passages in $D$. Formally, the task requires a function $f: (E_{\text{orig}}, D) \rightarrow (E_{\text{rev}}, P)$ that performs two subtasks: (1) extracting explicit modification directives from the documentary text, and (2) faithfully implementing those directives into revised criteria while preserving unmodified elements.

Protocol amendment documents provide naturally occurring ground-truth instances: each yields a triplet of original criteria, evidence document, and revised criteria. These documents vary in length, structure, and detail across institutions, phases, and therapeutic areas, exhibiting the same variation encountered across published papers, guidelines, and institutional reports. Validating on protocol documents therefore tests the core capability required for any evidence source. This framing is supported by prevalence: 76\% of protocols undergo at least one amendment, and 53\% modify eligibility criteria~\cite{getz2016amendments, getz2024benchmarks}.

Revised criteria are evaluated using an LLM-as-judge protocol. GPT-4o has demonstrated strong agreement with human evaluators on text equivalence and quality assessment tasks~\cite{zheng2024judging}. The judge (GPT-4o, accessed through UCSF Versa) scores each generated revision on a 0--4 ordinal scale: 0 indicates no overlap with the target revision, 1 indicates minimal overlap, 2 indicates partial overlap with notable gaps, 3 indicates the revision is largely equivalent to the target, and 4 indicates the revision is fully equivalent. We treat the ordinal score as the \emph{primary} metric because validation against ten domain experts (Section~\ref{sec:human_eval}) produces quadratic-weighted Cohen's $\kappa_{qw} = 0.694 \pm 0.096$ (``substantial'' per Landis \& Koch~\cite{landis1977measurement}), within the inter-human range $[0.531, 0.823]$; on the direct binary acceptability scale, $\kappa = 0.292 \pm 0.286$ (``fair''), with roughly triple the standard deviation. We report direct binary acceptability, the proportion scoring $\geq 3$, and normalized ordinal score (judge score divided by~4) alongside the primary ordinal metric. Full prompt templates and scoring rubrics are provided in Appendix~\ref{app:prompts}.

\section{RECITE System}
\label{sec:system}

\subsection{Framework Overview}

RECITE operates through a two-stage pipeline. In Stage~1 (Discovery), the system identifies relevant documentary evidence for a given trial's eligibility criteria by searching published literature, matching papers to trials, and extracting modification directives. In Stage~2 (Revision), given original eligibility criteria and discovered (or provided) evidence, the system generates revised criteria with provenance links to specific evidence passages. For the benchmark evaluation in Section~\ref{sec:benchmark}, Stage~1 is bypassed because the evidence documents are provided directly. For the real-world applications in Section~\ref{sec:applications}, Stage~1 employs the agentic literature discovery system described below.

The revision stage supports two modes. In truncated-to-context-window mode, the entire evidence document is provided as context alongside the original eligibility criteria. In RAG mode, the evidence document is first chunked and embedded using BAAI/bge-small-en-v1.5~\cite{xiao2024cpack} via LlamaIndex, and relevant passages are retrieved via dense passage retrieval with a top-$k$ of 10. Token budgets are enforced using tiktoken's cl100k\_base encoding: the query is truncated to 8,192 tokens before embedding, and retrieved passages are fitted within the LLM context window (16,384 tokens for locally hosted models) via LlamaIndex's compact response mode. Both modes produce revised eligibility criteria that cite specific passages as provenance for each change. This dual-mode design allows full-context processing of shorter evidence documents (such as targeted amendment rationale sections), while RAG mode supports aggregation across longer, multi-source evidence collections such as research papers and clinical guidelines.

\subsection{Agentic Literature Discovery}
\label{sec:discovery}

The literature discovery system uses orchestrator-driven iterative query expansion. A Python control loop alternates between LLM-generated search query formulation, adapter-based API search, and LLM-based relevance scoring. The LLM handles only the generative steps; all orchestration logic, rate limiting, and state management reside in the Python controller.

The system interfaces with two external APIs through dedicated adapters. The PubMed adapter issues boolean queries via NCBI E-utilities (esearch for identifier retrieval, efetch for metadata) with exponential backoff on transient failures. The Semantic Scholar adapter queries the Graph API, performing citation graph traversal to retrieve both the references and citing papers of each seed (up to 20 in each direction).

In each iteration, the system draws from a seed queue of high-relevance papers managed as a FIFO. If a seed is available, the LLM generates targeted queries informed by the seed's content; otherwise, queries are generated from scratch, avoiding duplicates tracked in a seen-identifier set. Queries are dispatched to PubMed and Semantic Scholar adapters. Each candidate paper is scored on relevance (1--5), extraction confidence (1--5), and accrual impact ease (1--5); papers above the relevance threshold enter the seed queue for progressive scope refinement. The orchestrator terminates after 5 consecutive rounds in which zero new relevant papers are discovered and the seed queue is empty.

For paper-trial matching, each paper passing directive extraction (relevance $\geq$ 4) yields extracted change directives and accrual impact estimates. The LLM generates ClinicalTrials.gov search queries targeting the public API (v2), and each (paper, trial) pair is scored for match quality and applicability (both 1--5). High-quality matches (both $\geq$ 4) seed additional queries to expand coverage iteratively. Full scoring prompts and rubrics are provided in Appendix~\ref{app:prompts}.

\subsection{Evidence-Grounded Revision}

The revision pipeline accepts original criteria and evidence and produces revised criteria with passage-level provenance. In RAG mode, evidence documents are chunked, embedded via dense passage retrieval~\cite{karpukhin2020dpr}, and the top-$k$ passages are provided as context. In truncated-to-context-window mode, the entire document serves as context. Both modes instruct the model to extract explicit directives, apply them, and cite supporting passages.

Directive extraction is validated by fuzzy matching against source text to prevent hallucination. The matching procedure uses a sliding-window approach over the evidence document, comparing each window to the extracted directive with SequenceMatcher-based similarity. Directives with low similarity scores are flagged as potentially unsupported; final validation was performed by manual review against source text.

\section{RECITE Benchmark}
\label{sec:benchmark}

Protocol amendment documents anchor the RECITE benchmark because each amendment produces a naturally occurring triplet---original eligibility criteria, documentary evidence, and revised criteria, requiring no manual annotation. These documents vary considerably in length, structure, and detail across sponsoring institutions, clinical phases, and therapeutic areas, similar to the diversity encountered in published papers, clinical guidelines, and institutional reports. Validating directive extraction on this heterogeneous evidence base therefore tests the core capability required for any documentary source type.

\subsection{Data Collection and Curation}

The dataset construction begins with 567,815 registered trials from the ClinicalTrials.gov bulk XML download. To identify trials with eligibility criteria modifications, a two-stage heuristic is applied. First, the ClinicalTrials.gov history API reports which modules changed in each version via \texttt{moduleLabels} metadata; only versions flagged as modifying the ``Eligibility'' module are retained. This reduces the candidate set from 567K to 59,996 trials without requiring full text comparison across all version pairs. Second, within this subset, both versions of the eligibility criteria text are normalized (whitespace collapsed, empty lines removed, HTML tags stripped) and compared using a \texttt{difflib}-based similarity check. Pairs with similarity above 0.80 or with fewer than 5\% of lines changed are discarded as metadata-only or formatting-only differences. Comparing consecutive version pairs across the retained trials produced 85,083 confirmed eligibility criteria changes across 49,369 unique trials.

To include grounding evidence, trials with downloadable protocol amendment PDFs were identified. The ClinicalTrials.gov API was queried for each of the 49,369 trials with EC changes, checking the largeDocumentModule metadata for protocol availability. Roughly 3,500 trials (approximately 7\%) had such documents. Evidence text was extracted from these PDFs using PyMuPDF, then joined with the corresponding eligibility criteria changes to produce complete triplets: original criteria, evidence document, and revised criteria. Amendment-to-version matching uses a date-based heuristic with a 7-day tolerance window; where dates are unavailable, a fallback assumes amendment $N$ corresponds to version $N$. The resulting dataset contains 11,913 instances across 5,735 trials. A stratified sample of 3,116 instances was selected for benchmarking to manage computational costs: evaluating all fifteen model configurations under both truncated-to-context-window and RAG conditions requires substantial resources given the mean evidence document length of approximately 134,000 characters. Because evidence documents frequently exceed model context windows, we report \emph{evidence retention} (Ret\%) for each context size: 8K models (Gemma~2) retain 23.5\% of the evidence text (12.5\% full-EC coverage); 32K models (Qwen~2.5 base) retain 81.1\% (53.5\%); and 128K models (GPT-4o, GPT-4o-mini, Qwen~2.5-7B-128K, Llama-3.1-70B, Qwen3-32B) retain 99.4\% (98.6\%). Each instance provides three components: (1) the original eligibility criteria prior to amendment, (2) grounding evidence extracted from the protocol amendment document, and (3) the target revised eligibility criteria as implemented in the subsequent protocol version.

\begin{table}[!htbp]
\caption{RECITE benchmark dataset summary.}
\label{tab:dataset_and_matches}
\centering
\footnotesize
\begin{tabular}{l@{\hskip 8pt}r}
\toprule
\textbf{Statistic} & \textbf{Value} \\
\midrule
Total instances & 11{,}913 \\
Unique trials & 5{,}735 \\
Sampled for benchmark & 3{,}116 \\
\midrule
Avg pre-amendment EC len.\ (chars) & 3{,}331 \\
Avg amended EC len.\ (chars) & 3{,}473 \\
Avg evidence len.\ (chars) & 134{,}180 \\
\midrule
\multicolumn{2}{l}{\textit{Trial phases}} \\
\quad Phase 1 / 1--2 / 2 & 433 / 439 / 1{,}663 \\
\quad Phase 3 / 4 & 1{,}059 / 316 \\
\quad Other / N/A & 451 / 1{,}374 \\
\midrule
Interventional / observational & 5{,}462 / 272 \\
Trial start date range & 2008--2023 \\
\midrule
\multicolumn{2}{l}{\textit{Top conditions (unique trials)}} \\
\quad Breast cancer & 98 \\
\quad Prostate cancer & 57 \\
\quad COVID-19 & 56 \\
\quad Obesity / Heart failure & 53 / 50 \\
\bottomrule
\end{tabular}
\end{table}

\subsection{Dataset Characteristics}

Table~\ref{tab:dataset_and_matches} summarizes the benchmark dataset. The 5,735 unique trials span a broad range of conditions, with breast cancer (98 trials), prostate cancer (57), COVID-19 (56), obesity (53), and heart failure (50) as the most represented. Phase~2 trials account for the largest share (1,663 trials), followed by trials with no reported phase (1,374), Phase~3 (1,059), and Phase~1/2 (439). Trial start dates range from 2008 to 2023. The average evidence document length is approximately 134,000 characters, with wide variation (up to 3.1M characters), reflecting the diversity of protocol documents across institutions and trial phases.

Protocol amendment documents differ considerably in length, structure, and level of detail across sponsoring institutions, clinical phases, and therapeutic areas. Some amendments provide tabular summaries of changes with explicit rationale; others embed modification directives within narrative justification spanning multiple pages. This variation is similar to the diversity of evidence sources the framework must handle, since published papers differ in structure from clinical guidelines, which differ from institutional reports.

\subsection{Model Evaluation}
\label{sec:model_eval}

Fifteen model configurations were evaluated on the 3,116-instance sample under truncated-to-context-window conditions. The suite spans (i) base open-weight models (Qwen~2.5 0.5B/3B/7B; Qwen~2.5-7B-128K; Gemma~2 2B/9B); (ii) commercial API models accessed through UCSF Versa, a PHI-safe HIPAA-compliant Azure deployment (GPT-4o-mini, GPT-4o~\cite{openai2024gpt4o}, GPT-4.1~\cite{openai2025gpt41}); (iii) scaling open-weights (Gemma~2-27B~\cite{gemmateam2024gemma2}, Qwen~2.5-72B~\cite{qwen2025qwen25}, Llama-3.1-70B~\cite{grattafiori2024llama31}, Qwen3-32B~\cite{qwenteam2025qwen3}); and (iv) two finetuned variants (Qwen~2.5-3B and Qwen~2.5-0.5B with QLoRA~\cite{dettmers2023qlora} on 720 stratified samples drawn from 8,838 non-benchmark instances).

\paragraph{Reporting strategy.} We present the ordinal 0--4 score as the primary metric and report direct binary acceptability and the proportion scoring $\geq 3$ alongside it. This choice is based on judge-against-human inter-rater reliability (Section~\ref{sec:human_eval}): on the ordinal scale, quadratic-weighted Cohen's $\kappa_{qw} = 0.694 \pm 0.096$ (``substantial'' per Landis \& Koch~\cite{landis1977measurement}); on the direct binary acceptability scale, $\kappa = 0.292 \pm 0.286$ (``fair''), with standard deviation roughly triple that of the ordinal scale.

\paragraph{Substantial-change subset.} Many EC version pairs in the benchmark differ only by trivial edits (whitespace, single-token rewordings) that any model accepts. To test directive extraction more rigorously, we report a \emph{substantial-change} subset using Gestalt pattern-matching similarity (Python's \texttt{SequenceMatcher}) at threshold sim$\leq$0.95, which retains $n = 1{,}351$ pairs (43.4\% of the benchmark).

Table~\ref{tab:table1_benchmark_hierarchy_v2} presents results for all fifteen configurations, with both full-benchmark and substantial-change splits.

\begin{table*}[!htbp]
\caption{RECITE benchmark across 15 model configurations on the full sample ($n = 3{,}116$) and the substantial-change subset (sim$\leq$0.95, $n = 1{,}351$, of full).
Bin Acc = direct binary acceptability from LLM judge (0/1). Ord = ordinal 0--4 score (mean $\pm \sigma$). NormOrd = ordinal/4. \%$\geq$3 / \%$\geq$4 = proportion of ordinal ratings at or above threshold.
Ret\% = evidence retained after truncation to model context window.
Reliability against ten domain experts: $\kappa_{\text{binary}} = .292 \pm .286$; $\kappa_{qw} = .694 \pm .096$ (Landis \& Koch~\cite{landis1977measurement}: ``fair'' vs.\ ``substantial'').
Finetuned: QLoRA~\cite{dettmers2023qlora} on 720 stratified samples drawn from 8{,}838 non-benchmark instances.}
\label{tab:table1_benchmark_hierarchy_v2}
\centering
\footnotesize
\setlength{\tabcolsep}{3pt}
\begin{tabular}{l rrr rr rr r}
\toprule
& \multicolumn{3}{c}{Aggregate} & \multicolumn{2}{c}{Full ($n{=}3{,}116$)} & \multicolumn{2}{c}{sim$\leq$.95 ($n{=}1{,}351$)} & \\
\cmidrule(lr){2-4} \cmidrule(lr){5-6} \cmidrule(lr){7-8}
Model & Bin Acc $\pm\sigma$ & Ord $\pm\sigma$ & NormOrd $\pm\sigma$ & \%$\geq$3 & \%$\geq$4 & \%$\geq$3 & \%$\geq$4 & Ret\% \\
\midrule
\multicolumn{9}{l}{\textit{Base Models}} \\
Qwen2.5-0.5B       & .01$\pm$.11 & 0.7$\pm$0.9 & .179$\pm$.22 &  3.8 &  1.4 &  3.2 &  1.0 & 81.1 \\
Qwen2.5-3B         & .09$\pm$.29 & 2.5$\pm$0.9 & .618$\pm$.23 & 49.5 & 12.6 & 42.8 &  8.0 & 81.1 \\
Qwen2.5-7B         & .18$\pm$.39 & 2.8$\pm$0.9 & .700$\pm$.22 & 63.1 & 22.7 & 56.3 & 16.9 & 81.1 \\
Qwen2.5-7B-128K    & .01$\pm$.08 & 0.2$\pm$0.6 & .050$\pm$.15 &  1.3 &  0.7 &  0.9 &  0.4 & 99.4 \\
Gemma2-2B          & .04$\pm$.20 & 1.9$\pm$0.9 & .483$\pm$.21 & 19.9 &  4.8 & 17.5 &  2.8 & 23.5 \\
Gemma2-9B          & .25$\pm$.44 & \textbf{3.0$\pm$0.9} & .742$\pm$.22 & 70.1 & \textbf{30.7} & 58.4 & 18.8 & 23.5 \\
\midrule
\multicolumn{9}{l}{\textit{API Models}} \\
GPT-4o-mini        & .18$\pm$.38 & 2.9$\pm$0.8 & .726$\pm$.19 & 69.7 & 22.4 & 63.4 & 17.4 & 99.4 \\
GPT-4o             & .20$\pm$.40 & \textbf{3.0$\pm$0.8} & .745$\pm$.19 & \textbf{75.3} & 25.0 & 68.4 & 19.0 & 99.4 \\
GPT-4.1            & .04$\pm$.19 & 2.7$\pm$0.6 & .673$\pm$.16 & 66.5 &  5.8 & 54.4 &  4.1 & 99.4 \\
\midrule
\multicolumn{9}{l}{\textit{Scaling Models}} \\
Gemma2-27B         & .23$\pm$.42 & 2.8$\pm$1.0 & .699$\pm$.25 & 63.7 & 27.1 & 54.0 & 16.7 & 23.5 \\
Qwen2.5-72B        & .23$\pm$.42 & \textbf{3.0$\pm$0.8} & .754$\pm$.20 & 77.0 & 27.8 & 67.6 & 18.1 & 97.3 \\
Llama-3.1-70B      & .38$\pm$.49 & 2.8$\pm$1.2 & .691$\pm$.31 & 61.6 & 37.2 & 52.6 & 28.6 & 99.4 \\
Qwen3-32B          & .35$\pm$.48 & 2.6$\pm$1.4 & .642$\pm$.36 & 60.8 & 34.6 & 54.1 & 25.9 & 99.4 \\
\midrule
\multicolumn{9}{l}{\textit{Finetuned Models (QLoRA, 720 stratified samples)}} \\
Qwen2.5-3B (QLoRA)  & .13$\pm$.34 & 2.6$\pm$0.9 & .645$\pm$.23 & 51.7 & 17.5 & 40.1 &  9.4 & 81.1 \\
Qwen2.5-0.5B (QLoRA)& .00$\pm$.06 & 0.5$\pm$0.8 & .130$\pm$.19 &  1.5 &  0.6 &  1.3 &  0.2 & 81.1 \\
\bottomrule
\end{tabular}
\end{table*}

\paragraph{Results.} On the full benchmark, GPT-4o achieves the highest ordinal score ($3.0 \pm 0.8$, normalized 0.745, 75.3\% scoring $\geq 3$); Gemma~2-9B and Qwen~2.5-72B match it within margin ($3.0 \pm 0.9$ and $3.0 \pm 0.8$ respectively). Qwen~2.5-7B-128K, despite its long-context capacity (Ret\%~99.4), achieves only $0.2 \pm 0.6$ ordinal; context length alone does not substitute for capability. On the substantial-change subset, all models show lower performance (e.g., GPT-4o drops from 75.3\% to 68.4\% scoring $\geq 3$), indicating that precise directive implementation remains an open problem even for the strongest models. Scaling within families is consistent: Qwen~2.5 (0.5B$\to$3B$\to$7B) shows ordinal $0.7 \to 2.5 \to 2.8$; Gemma~2 (2B$\to$9B$\to$27B) shows $1.9 \to 3.0 \to 2.8$. QLoRA finetuning of small models produces modest gains for the 3B (ordinal 2.5$\to$2.6) and minimal change for the 0.5B (0.7$\to$0.5).

RAG provides modest and variable gains across the long-context models tested in both conditions. For capable 128K-window models, the truncated-to-context-window setting already saturates available evidence (Ret\%~99.4); RAG gains primarily benefit small-context models on long evidence documents.

\begin{figure}[!htbp]
  \centering
  \includegraphics[width=0.6\textwidth]{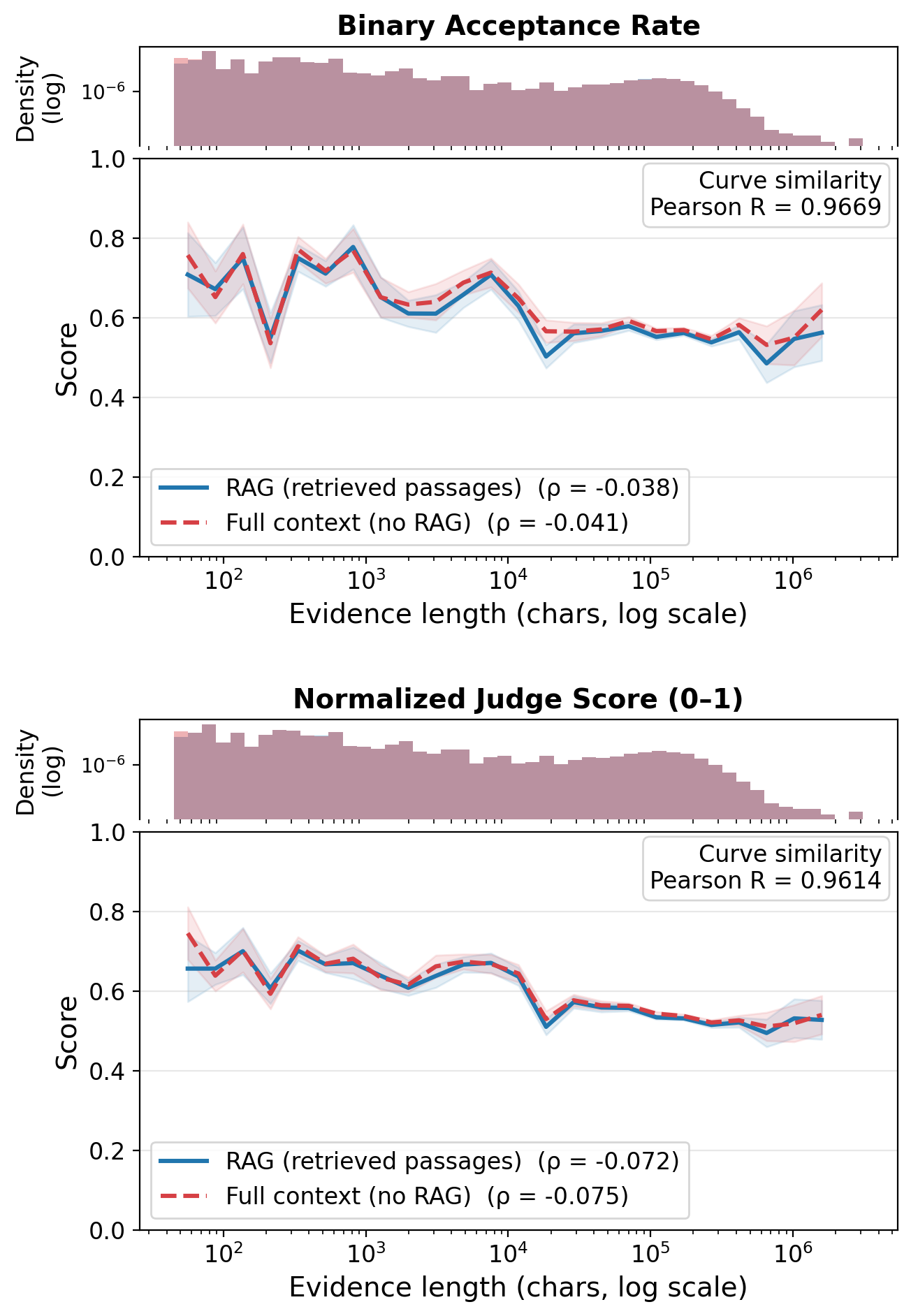}
  \caption{LLM judge performance as a function of evidence document length, pooled across the 128K-context models evaluated under both conditions. (a)~Direct binary acceptability rate. (b)~Normalized judge score (0--1). Each panel shows RAG and full-context (no-RAG) curves over 25 log-spaced length bins with 95\% CI bands; a marginal density histogram (log-scaled) along the top axis shows the evidence-length distribution.}
  \Description{Two vertically stacked panels showing direct binary acceptability and normalized judge score vs. evidence document length on a log scale, with RAG and no-RAG curves nearly overlapping across the full length spectrum.}
  \label{fig:evidence_length}
\end{figure}

Figure~\ref{fig:evidence_length} shows that evidence document length has negligible effect on revision quality (Spearman $\rho = -0.04$). The RAG and no-RAG curves are highly correlated (Pearson $R > 0.96$) and differ primarily in level rather than trend. In the single-document setting, even documents exceeding $10^6$ characters remain within capable models' processing range; retrieval-based methods will become necessary when extending to multi-document evidence sources.

\subsection{Human Evaluation of the Automated Judge}
\label{sec:human_eval}

To validate the GPT-4o-as-judge protocol, we conducted a structured human evaluation in which ten domain expert judges rated 160 held-out instances, producing 200 paired ratings against the automated judge. Inter-human reliability spans a moderate-to-substantial range, $\kappa_{qw} \in [0.531, 0.823]$. EC equivalence judgments retain irreducible expert subjectivity even with a fixed rubric.

Comparing the automated judge to the human ratings yields:
\begin{itemize}
\item \textbf{Ordinal (0--4)}: $\kappa_{qw} = 0.694 \pm 0.096$ --- ``substantial'' per Landis \& Koch~\cite{landis1977measurement}.
\item \textbf{Binary}: $\kappa = 0.292 \pm 0.286$ --- ``fair''. The standard deviation is roughly triple that of the ordinal scale; binary judgment discards information about partial agreement.
\end{itemize}
The ordinal $\kappa_{qw}$ falls within the inter-human range, so the automated judge is not measurably worse than a representative human rater on the 0--4 scale. We therefore retain ordinal as the primary metric throughout the paper and report direct binary acceptability for continuity with prior work.

\section{Real-World Applications}
\label{sec:applications}

\begin{table*}[!htbp]
\caption{All paper--trial matches with extracted directives and accrual impact quotes.
\textbf{Des/EC}: source paper provides trial-design recommendations / explicit EC modification directives (T/F).
ECOG GU and ECOG all reclassified F/F (document scoring practices, not protocol-level EC changes).
Enrollment columns show current and directional estimate under broadened criteria; estimates are omitted (\textemdash) for F/F papers whose reported gains reflect scoring-practice changes, not protocol-level EC modifications.
M/A: LLM match\,/\,applicability scores (1--5).}
\label{tab:accrual_papers_quotes}
\centering
\footnotesize
\setlength{\tabcolsep}{2.5pt}
\begin{tabular}{@{} p{1.5cm} c p{3.2cm} p{3.2cm} l r @{\;$\rightarrow$\;} r c @{}}
\toprule
Source & Des/EC & EC Change Directive & Accrual Impact Quote & Trial (NCT ID) & \multicolumn{2}{c}{Enrollment} & M\,/\,A \\
\cmidrule(lr){6-7}
 & & & & & Current & \multicolumn{1}{c}{Est.} & \\
\midrule
\multirow{2}{=}{\raggedright\textbf{Harvey}~\cite{kaur2021broadening}}
 & \multirow{2}{*}{T/T}
 & \multirow{2}{=}{\raggedright\footnotesize Allow brain mets, other malignancies, CrCl $\geq$30\,mL/min}
 & \multirow{2}{=}{\raggedright\footnotesize Potential to double eligible patient population}
 & NCT03101995 & 18 & 36 & 4/5 \\
 & & & & NCT04163432 & 43 & 86 & 4/5 \\[0.4em]
 \rule{0pt}{0pt} & & & & & \multicolumn{2}{c}{} & \\   
\addlinespace[0.4em]\midrule
\multirow{8}{=}{\raggedright\textbf{Kim}~\cite{kim2019broadening}}
 & \multirow{8}{*}{T/T}
 & \multirow{8}{=}{\raggedright\footnotesize Allow brain mets, other malignancies, lower CrCl to $>$30\,mL/min}
 & \multirow{8}{=}{\raggedright\footnotesize Nearly 2$\times$ aNSCLC pts eligible (4851, 46.2\%)}
 & NCT02792582 & 160 & 320 & 3/4 \\
 & & & & NCT03367819 & 44 & 88 & 3/4 \\
 & & & & NCT04114136 & 72 & 144 & 4/5 \\
 & & & & NCT04488874 & 50 & 100 & 3/4 \\
 & & & & NCT04739761 & 506 & 1{,}012 & 5/5 \\
 & & & & NCT05588453 & 30 & 60 & 4/5 \\
 & & & & NCT06157541 & 58 & 116 & 4/5 \\
 & & & & NCT06481813 & 730 & 1{,}460 & 4/5 \\
\addlinespace[0.4em]\midrule
\multirow{6}{=}{\raggedright\textbf{KRAS G12C}~\cite{kras2025multicenter}}
 & \multirow{6}{*}{T/T}
 & \multirow{6}{=}{\raggedright\footnotesize Liberalize renal dysfunction and brain mets criteria}
 & \multirow{6}{=}{\raggedright\footnotesize +25\% enrollment without decreasing median OS}
 & NCT01509937 & 433 & 541 & 3/3 \\
 & & & & NCT04013620 & 48 & 60 & 4/4 \\
 & & & & NCT04467021 & 61 & 76 & 4/4 \\
 & & & & NCT04704258 & 100 & 125 & 3/4 \\
 & & & & NCT05534620 & 36 & 45 & 4/4 \\
 & & & & NCT06531824 & 11{,}000 & 13{,}750 & 4/4 \\
\addlinespace[0.4em]\midrule
\multirow{10}{=}{\raggedright\textbf{Fung}~\cite{hcc2019eligibility}}
 & \multirow{10}{*}{T/T}
 & \multirow{10}{=}{\raggedright\footnotesize Include ECOG~2 or Child-Pugh B7 patients}
 & \multirow{10}{=}{\raggedright\footnotesize $>$2$\times$ patients treated if MEC applied}
 & NCT00485303 & 58 & 141 & 4/5 \\
 & & & & NCT01273155 & 72 & 175 & 5/5 \\
 & & & & NCT01967576 & 14 & 34 & 4/5 \\
 & & & & NCT02314143 & 48 & 116 & 3/4 \\
 & & & & NCT03494816 & 24 & 58 & 4/4 \\
 & & & & NCT06889675 & 36 & 87 & 4/5 \\
 & & & & NCT06926673 & 62 & 151 & 4/4 \\
 & & & & NCT06984497 & 32 & 78 & 4/4 \\
 & & & & NCT06990659 & 206 & 500 & 4/4 \\
 & & & & NCT07037394 & 252 & 612 & 4/4 \\
\addlinespace[0.4em]\midrule
\multirow{10}{=}{\raggedright\textbf{ECOG GU}~\cite{ecog2023genitourinary}}
 & \multirow{10}{*}{F/F}
 & \multirow{10}{=}{\raggedright\footnotesize Edit ECOG 2$\to$1 (prostate), 3$\to$2 (bladder)}
 & \multirow{10}{=}{\raggedright\footnotesize +233\% (prostate), +300\% (bladder) matched trials/pt}
 & NCT00428714 & 73 & \textemdash & 4/4 \\
 & & & & NCT00485303 & 58 & \textemdash & 4/5 \\
 & & & & NCT00699751 & 921 & \textemdash & 4/5 \\
 & & & & NCT01756638 & 48 & \textemdash & 4/5 \\
 & & & & NCT03768570 & 82 & \textemdash & 4/4 \\
 & & & & NCT04960709 & 712 & \textemdash & 4/4 \\
 & & & & NCT05171387 & 78 & \textemdash & 4/5 \\
 & & & & NCT05348876 & 50 & \textemdash & 4/5 \\
 & & & & NCT06524544 & 320 & \textemdash & 4/4 \\
 & & & & NCT06661122 & 100 & \textemdash & 4/4 \\
\addlinespace[0.4em]\midrule
\multirow{8}{=}{\raggedright\textbf{ECOG all}~\cite{ecog2023profound}}
 & \multirow{8}{*}{F/F}
 & \multirow{8}{=}{\raggedright\footnotesize Edit ECOG 3$\to$2 or 2$\to$1 across cancer types}
 & \multirow{8}{=}{\raggedright\footnotesize +246\% and +189\% matched CTs ($p{<}0.001$)}
 & NCT00022672 & 208 & \textemdash & 3/3 \\
 & & & & NCT00485303 & 58 & \textemdash & 4/5 \\
 & & & & NCT01967576 & 14 & \textemdash & 4/5 \\
 & & & & NCT03022032 & 102 & \textemdash & 3/4 \\
 & & & & NCT03288987 & 126 & \textemdash & 4/5 \\
 & & & & NCT03355079 & 2 & \textemdash & 4/5 \\
 & & & & NCT05721755 & 290 & \textemdash & 4/4 \\
 & & & & NCT06677255 & 149 & \textemdash & 3/4 \\
\midrule
\multicolumn{8}{@{}l}{\textbf{Total: 6 source papers, 44 paper--trial matches; Des=T/EC Mod=T for 4 papers; reclassified F/F for 2 (ECOG GU, ECOG all).}} \\
\bottomrule
\end{tabular}
\end{table*}

To evaluate RECITE beyond the protocol-based benchmark, we applied the full two-stage pipeline to published research literature. The agentic crawling system (Section~\ref{sec:discovery}) retrieved papers from PubMed that recommend specific eligibility criteria changes, matched them to registered trials on ClinicalTrials.gov, and extracted modification directives with associated accrual impact estimates.

\subsection{Literature--Trial Matching}

The system identified 44 paper--trial matches across 6 source publications and 41 unique trials, spanning ASCO-Friends criteria broadening~\cite{kaur2021broadening, kim2019broadening}, organ function threshold liberalization~\cite{kras2025multicenter}, expanded hepatocellular carcinoma criteria~\cite{hcc2019eligibility}, and ECOG performance status adjustments~\cite{ecog2023genitourinary, ecog2023profound}. Each match is a pairing in which the paper's recommended eligibility change is applicable to the trial's current criteria, as scored by the LLM pipeline described in Section~\ref{sec:discovery}.

Table~\ref{tab:accrual_papers_quotes} presents all matches grouped by source paper with current and estimated enrollment. Directives range from targeted threshold changes to broader performance status adjustments, all validated by fuzzy matching against source publications. Match quality scores and rubrics are in Appendix~\ref{app:full_matches} and Appendix~\ref{app:prompts}.

\subsection{Estimated Accrual Impact}
\label{sec:source_evidence}

Here, we present the underlying quoted evidence for the estimated accrual impact of each study, when possible. We also provide a binary classification of each source for whether it (i) provides trial-design recommendations (Des) and (ii) issues explicit directives to modify eligibility criteria (EC Mod):
\begin{itemize}
\item \textbf{Harvey et al.}~\cite{kaur2021broadening} [Des: T, EC Mod: T] --- ``Broadening [brain metastases, prior malignancies, CrCl $\geq$ 30~mL/min] has the potential to \emph{double} the eligible patient population.''
\item \textbf{Harvey et al.}~\cite{kim2019broadening} [Des: T, EC Mod: T] --- ``Patients with treated brain metastases should be routinely included''; ``strive for inclusiveness'' (ASCO-Friends consensus).
\item \textbf{KRAS G12C}~\cite{kras2025multicenter} [Des: T, EC Mod: T] --- ``Liberalizing renal dysfunction and brain metastases criteria yielded +25\% enrollment without decreasing median OS.''
\item \textbf{Fung et al.}~\cite{hcc2019eligibility} [Des: T, EC Mod: T] --- ``Expanding to [ECOG PS 2 or Child-Pugh B7] increased eligibility to 31.7\%'' (from 13.1\%).
\end{itemize}
All four source papers issue explicit directives to broaden eligibility. The directional signal is consistent: liberalization of brain-metastasis exclusions, prior-malignancy exclusions, organ-function thresholds, and ECOG cutoffs is associated with substantially expanded eligibility in the source-paper analyses. We do not assert specific percentage gains for the matched trials; downstream verification against actual enrollment outcomes remains future work.

\subsection{Scope and Interpretation}

The current application focuses on oncology, given the maturity of EC liberalization research in that area; extension to other therapeutic areas is architecturally straightforward but not yet empirically validated. All matches were among the highest-scoring pairs (see Appendix~\ref{app:prompts}).

\subsection{Illustrative Example}

To illustrate how a paper's directive applies to a specific trial, consider Harvey et al.~\cite{kaur2021broadening} matched to NCT04163432 (Durvalumab chemo-immunotherapy for metastatic non-squamous NSCLC; match 4/5, applicability 5/5). The extracted directive recommends allowing patients with brain metastases, other malignancies, and a creatinine clearance of $\geq$\,30\,mL/min. The original trial already included a renal criterion of eGFR $\geq$\,30\,mL/min, consistent with the paper's CrCl recommendation. However, the trial excluded patients with ``active central nervous system metastases and/or carcinomatous meningitis'' and ``other active malignancy within 2 years except for basal cell carcinoma of the skin or carcinoma in situ of the cervix.'' Applying the directive broadens both exclusions to permit controlled brain metastases and prior malignancies beyond the narrow exceptions. The source paper reports that broadening these three criteria has ``the potential to double the eligible patient population''~\cite{kaur2021broadening}. Additional before-and-after criteria examples are included in Appendix~\ref{app:directive_examples}.

\section{Limitations and Ethical Considerations}
\label{sec:limitations}

\paragraph{Data and benchmark.}
The RECITE benchmark draws from a specific subset of the trial population: of 567,815 registered trials, only those with publicly downloadable protocol amendment PDFs contribute evidence documents, producing roughly 3,500 trials (0.6\% of the registry). Table~\ref{tab:dataset_and_matches} shows that Phase~2 and Phase~3 trials account for a disproportionate share, and top conditions are weighted toward oncology and well-funded disease areas. The benchmark construction pipeline relies on two key heuristics that trade completeness for efficiency. First, eligibility criteria changes are identified via ClinicalTrials.gov \texttt{moduleLabels} metadata rather than exhaustive version-pair comparison across all 567K trials, potentially missing changes where registry metadata is incomplete. Second, text-level change detection uses similarity thresholds that filter out very small substantive changes alongside formatting-only differences. The benchmark validates the framework on protocol documents, and the applications in Section~\ref{sec:applications} demonstrate transfer to published papers; systematic evaluation across the full range of document types (e.g., clinical guidelines, institutional reports) remains future work. The LLM-as-judge evaluation was validated through manual inspection of scored instances. The equivalence-judging task (determining whether two eligibility criteria texts are functionally identical) is well-suited to LLM evaluation given its reliance on structured text comparison. However, LLM judges may exhibit systematic biases related to hallucination or scoring precision, and inter-rater reliability against human expert judgment has not been measured at scale.

\paragraph{Applications.}
The 44 paper--trial matches represent the highest-scoring pairs within the scoring system (see Appendix~\ref{app:prompts}), not a random sample. Reported accrual signals are inherited directly from the source publications and treated as directional rather than as independent numeric predictions for the matched trials (Section~\ref{sec:source_evidence}). RECITE surfaces and operationalizes existing research conclusions rather than generating enrollment forecasts. Mechanically scaling source-paper percentages onto matched trials would assume patient-population homogeneity that the source papers do not establish. These estimates were not independently verified against actual enrollment data. The current applications focus on oncology, and directive extraction operates on abstracts rather than full text; more detailed recommendations in methods or results sections may therefore be missed.

\paragraph{Ethics.}
RECITE is a decision-support tool, not a replacement for clinical judgment. All proposed revisions require review by clinical investigators, IRB approval, and regulatory review where applicable~\cite{fda2020diversity}. The system provides provenance linking each revision to specific source passages to support expert evaluation. The published literature from which RECITE draws recommendations may itself reflect biases in what has been studied. Although broadening eligibility criteria generally improves demographic diversity~\cite{kaur2024broadening}, RECITE does not explicitly optimize for representativeness.

\paragraph{Future work.}
Several directions remain. Extending directive extraction to full paper text (beyond abstracts) would capture more detailed recommendations in methods and results sections. Constructing a literature-linked benchmark from protocol document references would complement our protocol-based benchmark with a paper-based evaluation setting. Matching modified criteria to actual enrollment outcomes using longitudinal registry data would provide ground-truth validation of the accrual impact estimation pipeline. Temporal meta-analysis of how eligibility criteria evolve in response to published evidence would further clarify the relationship between evidence publication and protocol modification. Finally, assessing and improving the robustness of LLM-inferred directives through tests specifically tailored to trial design specifications~\cite{xian2025robustness} would strengthen the trustworthiness of extracted directives and represents a natural continuation of this line of work.

\section{Conclusion}
\label{sec:conclusion}

RECITE provides infrastructure to discover, extract, and implement evidence-based eligibility criteria revisions with full provenance. On a benchmark of 3,116 instances derived from protocol amendments on ClinicalTrials.gov, the best configurations (GPT-4o, Gemma~2-9B, Qwen~2.5-72B) achieve ordinal scores of approximately 3.0/4 under an LLM-as-judge protocol validated against ten domain experts ($\kappa_{qw} = 0.694$, ``substantial''). Applied to published literature through an agentic discovery system, RECITE identifies 44 paper--trial matches across five publications providing evidence-grounded directives for EC liberalization. Both evidence types (protocol amendments and published papers) share the property that makes this task tractable: explicit modification directives embedded in varied text. The underlying approach of extracting directives and implementing them in design documents could generalize to other trial elements such as endpoint selection or dosing regimens.

\section*{GenAI Disclosure}

\paragraph{Research methodology.}
Large language models are the primary subject of this work. Fifteen LLM configurations were evaluated across base, scaling, API, and finetuned (QLoRA) categories. GPT-4o served as the LLM-as-judge evaluator (Section~\ref{sec:formulation}), validated against ten domain experts. LLMs were also used within the agentic literature discovery pipeline for search query formulation, relevance scoring, and directive extraction (Section~\ref{sec:discovery}).

\paragraph{Manuscript preparation.}
Claude (Anthropic) was used to assist with editing, restructuring, and refining manuscript prose. All generated or modified text was reviewed, verified, and approved by the authors.

\paragraph{Code development.}
Claude (Anthropic) was used for code completion and debugging support. All code was reviewed and tested by the authors.

\subsubsection*{Acknowledgments}
This work was supported in part by the National Science Foundation Graduate Research Fellowship Program. Computational resources were provided by UCSF Versa. The RECITE framework is available at \url{https://github.com/abbasilab/RECITE}.

\bibliographystyle{plainnat}
\bibliography{references}

\FloatBarrier
\appendix


\section{Full Paper--Trial Matches}
\label{app:full_matches}

All 44 paper--trial matches with enrollment impact estimates, source papers, extracted directives, and LLM evaluation scores are presented in Table~\ref{tab:accrual_papers_quotes}. For each match, the Match score (M, 1--5) assesses topical alignment between the paper's recommendations and the trial's eligibility criteria, and the Applicability score (A, 1--5) measures how directly the paper's directive can be implemented in the trial. Scoring rubrics and prompt templates are provided in Appendix~\ref{app:prompts}.

\section{Scoring Prompts and Rubrics}
\label{app:prompts}

This appendix documents the LLM scoring prompts, rubrics, thresholds, and filtering logic used throughout the RECITE pipeline. All prompts are stored in JSON configuration files and loaded at runtime. Full prompt templates are available in the supplementary materials.

\subsection{Paper Relevance Scoring}
\label{app:paper_scoring}

Each retrieved paper is scored on three dimensions (1--5 each). Table~\ref{tab:scoring_rubrics} shows the relevance rubric (used for initial filtering) alongside the match and applicability rubrics (used for paper--trial pairing). Extraction confidence measures how actionable the paper's recommendations are (1 = no actionable information; 5 = directly applicable threshold changes). Accrual ease measures how quantifiable the enrollment impact is (1 = vague or subjective; 5 = easily quantifiable biomarkers). Papers with relevance $\geq 3$ proceed to directive extraction.

\begin{table}[!htbp]
\caption{Scoring rubrics for paper relevance (left), paper--trial match alignment (center), and applicability (right). Relevance is scored per-paper; match and applicability are scored per (paper, trial) pair.}
\label{tab:scoring_rubrics}
\small
\begin{tabular}{cp{0.22\columnwidth}p{0.22\columnwidth}p{0.22\columnwidth}}
\toprule
Score & Relevance & Match & Applicability \\
\midrule
1 & Not relevant to EC or enrollment & No alignment; different conditions & Change implied, requires deep digging \\
2 & Tangentially relevant; mentions trials broadly & Minimal; broad topic overlap only & Vague hints; may require full paper \\
3 & Discusses EC but not central focus & Some alignment; similar condition, partial EC overlap & Mentioned but not overt; some interpretation needed \\
4 & Focuses on EC, enrollment barriers, or participation & Good alignment; same/similar condition, substantial overlap & Directive present, minor interpretation needed \\
5 & Directly about EC modification or broadening & Strong alignment across all dimensions & Overt in abstract; immediately applicable \\
\bottomrule
\end{tabular}
\end{table}

The scorer also extracts verbatim \textbf{change directive quotes} and \textbf{change rationale quotes}. High-quality matches (both scores $\geq 4$) seed additional query generation. Default filtering: match $\geq 3$, sorted by match then applicability descending, maximum 10 matches per paper.

\subsection{LLM Judge (Benchmark Evaluation)}
\label{app:judge}

The benchmark judge scores each (target, predicted) eligibility criteria pair on two scales: a \textbf{binary score} (0 = incorrect, 1 = correct) and an \textbf{ordinal score} on a 0--4 scale where 0 indicates no correspondence with the target, 1 indicates poor correspondence with major omissions, 2 indicates partial correspondence with notable gaps, 3 indicates good correspondence with minor differences, and 4 indicates the prediction is essentially identical to the target. The judge returns both scores independently; Bin Acc in the main table reflects the direct binary judgment, while \%$\geq$3 reports the proportion of ordinal scores at or above~3. For efficiency, predictions are evaluated in batches, with the judge returning a JSON dictionary mapping instance indices to score pairs.

\subsection{Directive and Impact Extraction}
\label{app:directives}

The directive extraction prompt classifies each paper as containing exact directives (extractable from abstract), directives requiring full text, or no directives. Only papers with extractable directives proceed. The impact extraction prompt identifies numeric accrual estimates (percent increase or absolute patient counts) with supporting verbatim evidence. Key computed quantities are: \texttt{paper\_pct\_gain} (percent increase in eligible patients, e.g., 100 = doubling) and \texttt{scalar\_gain} (estimated additional patients for a given trial = enrollment $\times$ paper\_pct\_gain / 100).

\subsection{Query Generation}
\label{app:queries}

Three query generation prompts are used: (1)~literature search queries for PubMed and Semantic Scholar, (2)~trial matching queries for ClinicalTrials.gov derived from paper content, and (3)~seed expansion queries generated from high-quality matches. All prompts receive previously issued queries to avoid repetition and return JSON arrays of query strings.

\section{Representative Directive Applications}
\label{app:directive_examples}

Six examples illustrate how RECITE applies paper directives to matched trial eligibility criteria. Examples~1, 3, 4, 5 are actionable broadening directives (T/T in Table~\ref{tab:accrual_papers_quotes}); Examples~2 and~6 (both Tsaur et al.) are F/F scoring-practice observations. The full set of 44 applications is available in the supplementary materials.

\subsection{Example 1: ASCO-Friends Broadening (Durvalumab Trial)}

\noindent\textbf{Source:} Harvey et al.~\cite{kaur2021broadening} (Match 4/5, Applicability 5/5) \quad \textbf{Trial:} NCT04163432 (Durvalumab, metastatic non-squamous NSCLC) \\
\textbf{Directive:} Allow patients with brain metastases, other malignancies, and CrCl $\geq$30\,mL/min. \quad \textbf{Accrual gain:} +100\% (43 patients)

The trial's renal criterion (eGFR $\geq$30) already satisfies the CrCl recommendation. The primary broadening removes exclusions for controlled brain metastases and prior non-lung malignancies.

\subsection{Example 2: ECOG Scoring-Practice Observation (F/F, Prostate Cancer)}

\noindent\textbf{Source:} Tsaur et al.~\cite{ecog2023genitourinary} (Match 4/5, Applicability 5/5) \quad \textbf{Trial:} NCT00485303 (Abiraterone, mCRPC) \\
\textbf{Directive (extracted):} Edit ECOG from 2 to 1 for prostate cancer patients. \quad \textbf{Classification:} F/F

The source documents inter-rater ECOG scoring variability, not a protocol-level EC modification. Re-scoring ECOG~2$\to$1 would expand matched trials (+233\%) but does not constitute an actionable EC change.

\subsection{Example 3: ASCO-Friends Broadening (T-DXd HER2 Trial)}

\noindent\textbf{Source:} Harvey et al.~\cite{kim2019broadening} (Match 5/5, Applicability 5/5) \quad \textbf{Trial:} NCT04739761 (Trastuzumab deruxtecan, HER2+ breast cancer with brain metastases) \\
\textbf{Directive:} Allow brain metastases, other malignancies, and CrCl $>$30\,mL/min. \quad \textbf{Accrual gain:} +100\% (506 patients)

This trial already permits brain metastases (5/5 match). The primary broadening applies to the ``other malignancies'' exclusion and creatinine clearance threshold.

\subsection{Example 4: Renal and Brain Metastasis Liberalization (KRAS G12C Trial)}

\noindent\textbf{Source:} KRAS G12C multicenter study~\cite{kras2025multicenter} (Match 3/5, Applicability 3/5) \quad \textbf{Trial:} NCT01509937 (Phase I/II, renal dysfunction) \\
\textbf{Directive:} Liberalize criteria for renal dysfunction and brain metastases. \quad \textbf{Accrual gain:} +25\% (108 patients)

The moderate match (3/5) reflects a transplant rather than NSCLC context, but renal function threshold liberalization remains applicable. This illustrates cross-specialty applicability of broadening recommendations.

\subsection{Example 5: HCC Performance Status Expansion (Belinostat Trial)}

\noindent\textbf{Source:} Fung et al.~\cite{hcc2019eligibility} (Match 5/5, Applicability 5/5) \quad \textbf{Trial:} NCT01273155 (Belinostat, solid tumors/lymphomas, Phase I/II) \\
\textbf{Directive:} Include ECOG~2 patients and limited liver dysfunction (Child-Pugh B7). \quad \textbf{Accrual gain:} +143\% (103 patients)

The trial already allows ECOG $\leq$2. The primary broadening is expanding hepatic function criteria to include Child-Pugh B7 patients with limited liver dysfunction.

\subsection{Example 6: ECOG Scoring-Practice Observation (F/F, Pan-Cancer)}

\noindent\textbf{Source:} Tsaur et al.~\cite{ecog2023profound} (Match 4/5, Applicability 5/5) \quad \textbf{Trial:} NCT03288987 (Stivant, metastatic colorectal cancer) \\
\textbf{Directive (extracted):} Edit ECOG 3$\to$2 or 2$\to$1 to increase matched trials. \quad \textbf{Classification:} F/F

Pan-cancer companion to Example~2. Documents systematic ECOG overestimation (+246\% matched trials, $p<0.001$) rather than recommending protocol-level EC changes.

\end{document}